# A Comparative and Experimental Study on Automatic Question Answering Systems and its Robustness against Word Jumbling

**Haoran Hu**
haoranhu@umass.edu
**Sai Sameer Vennam**
svennam@umass.edu

**Shashidhar Reddy Javaji**
sjavaji@umass.edu
**Vijaya Gajanan Buddhavarapu**
vbuddhavarap@umass.edu

## 1 Problem statement

Question answer generation using Natural Language Processing models is ubiquitous in the world around us. It is used in many use cases such as the building of chat bots, suggestive prompts in google search and also as a way of navigating information in banking mobile applications etc. It is highly relevant because a frequently asked questions (FAQ) list can only have a finite amount of questions but a model which can perform question answer generation could be able to answer completely new questions that are within the scope of the data (Sheetal Rakangor, 2015). This helps us to be able to answer new questions accurately as long as it is a relevant question. In commercial applications, it can be used to increase customer satisfaction and ease of usage. However a lot of data is generated by humans so it is susceptible to human error and this can adversely affect the model's performance and we are investigating this through our work (Chidinma A. Nwafor, 2021).

### 1.1 Motivation

This project aims to test the robustness of an automatic question answering system when words in context are scrambled or deleted. Many times we contain large amounts of textual data which is not exactly of good quality. The textual data that is generated by a large volume of people usually has a lot of spelling mistakes and word distortions. This can be due to a lack of native speakers or it could also be due to different dialects using different spellings for the same word. There is a need for question answering systems to be able to handle all kinds of text which might have a lot of spelling mistakes and other errors too. There is a need to understand how well the state-of-the-art language models are performing and also how well these models can perform at different levels of noise in the data. We aim to understand and improve the performance of these state of the art models on different levels of noise. A model that is robust to context corruption can increase the satisfaction of the users of the system, especially for people whose English is not their native language. This model will still be robust enough to understand and answer questions appropriately. This is relevant because many times we do not have control of the sources of data and how it is generated. There might be a need to perform question-answering in real time or there could also be a time/resource constraint to pre-processing this text. In such cases, it becomes highly important for the models to be able to understand and process this text which is riddled with noise. The results of our study could help in choosing and training the appropriate models to perform this task.

### 1.2 Our project

In our project, we compare the performance degradation of four different NLP models when the datasets are corrupted in two distinct ways: Word jumbling and word deletion. There has not been work done that compares which kind of syntactic aberrations affect QA-systems more and which models can deal with the said aberrations.

### 1.3 Outline of this project

In Section 2, we compare what we proposed and what we accomplished at this time. In Section 3, we discuss related work on QA systems and how our project differ from theirs. In Section 4, we introduce our dataset: The Stanford Question and Answer dataset (SQUAD) (Rajpurkar et al., 2016), a frequently used dataset for Question Answering tasks. In Section 5, we introduce the baseline results of the four chosen NLP models and we discuss hyperparameters that were present during the experiments and which values we chose for said hyperparameters. We also discuss the evaluation metrics on which the baseline models are evaluated upon. In Section 6, we go over our proposed approach in detail. We also showcase the results from our experiments and compare them with the baseline results. In Section 7, we will perform error analysis on both the baseline and experimental results. In Section 8, we reflect upon our work in the conclusion. Finally, section 9 details each team member's contribution to this project.

## 2 What you proposed vs. what you accomplished

- **Import and use SQUAD dataset:** With the help of HuggingFace datasets, we were able to use SQUAD and preprocess it to introduce noise (in the form of word jumbling) in the 'context' of each sample.

- **Build baselines on BART and T5:** We were not able to accomplish this, even when using additional GPU resources provided by Google Colab Pro. T5-small took extremely long to train (12 hours for an epoch) and the notebook runtime ended before its completion. Facebook's BART-base, while smaller than BART-large, suffered from CUDA OOM (Out of memory) issues even with smaller batch sizes.

- **Used different models:** We used different models that were not in our proposal for our experiments: BERT, ALBERT, ROBERTA, and DistilBERT. After significant testing of different transformer models, we found that we could import and implement our experiments using the GPU resources that we had on these models. These models took at most 2 hours per epoch to train.

- **Type of study:** It is a comparative as well as an experimental study where we have experimented with different models and their performance when noise is added to the dataset, we have added the noise external which we have taken as a type of possible human error in the text. We were able to generate results for noise introduced in the form of word jumbling and word deletion at 3 different noise levels which were 5, 10, and 15 percent.

## 3 Related work

Question Answering for a question asked in context of a specific page script has been studied by many and show promising results. A pre-trained T5 For Conditional Generation base model is taken and fine-tuned on the question answering task using the SQuAD 2.0, The Stanford Question Answering Dataset(Li et al., 2019). An alternative approach to this is using the pre-trained DistilBERT model(Patil et al., 2022). (Du et al., 2017) introduce an attention-based sequence learning model for the QG task, and the question generated by their system are also rated as being more natural and as more difficult to answer. (Krishna and Iyyer, 2019) implemented a SQUASH (Specificity-controlled Question-Answer Hierarchies) pipeline to generate question-answer pairs, they use GPT-2 (Radford et al., 2018) to convert the input sentence into a question, and BERT (Devlin et al., 2018) is used to answer the question generated by GPT-2 as an output answer. (Alberti et al., 2019) use BERT to introduce a generating synthetic question answering corpora by combining models of question generation and answer extraction. (TSAI et al., 2021) proposes a combination of syntax-based and semantic-based Automatic Question Generation (AQG) system, using BERT for semantic analysis and Stanford CoreNLP (Manning et al., 2014) for syntax analysis, then construction question by GPT-2. Even though many Automatic Question Generation (AQG) system have shown promising results. None of them has been tested on whether they can survive from word scrambling or word deletion in the context. We use The Stanford Question and Answer dataset (SQUAD) and corrupted part of the dataset and finetuned several state of arts NLP models using the corrupted SQUAD. We then evaluated the performance of the models with corrupted validation set in SQUAD using several evaluation metrics.

## 4 Your dataset

The Stanford Question and Answer dataset (SQUAD) (Rajpurkar et al., 2016) is a reading comprehension dataset, consisting of questions posed by crowd workers on a set of Wikipedia articles. It contains approximately 150,000 example triplets (question, context, and answer). Some of the questions are unanswerable. There are two versions of the SQUAD dataset. The first iteration is SQUAD 1.1 which consists of 100,000 questions and in the latest iteration which is SQUAD 2.0, the 100,000 questions present in SQUAD 1.1 are combined with another 50,000 questions to form the SQUAD 2.0 dataset. Since we are only interested in the performance of the AQG system in answering the questions with the corrupted context, we use SQUAD1 instead of SQUAD2 for our project. Each example in the dataset contains id, title, context, question, and answer. The answer not only contains the answer to the question but also the location of the answer inside the context.

Sample from SQUAD dataset:
'id': '5733be284776f41900661182
'title': 'University of Notre Dame',
'question': 'To whom did the Virgin Mary allegedly appear in 1858 in Lourdes France?',
'context': 'Architecturally, the school has a Catholic character. Atop the Main Building's gold dome is a golden statue of the Virgin Mary. Immediately in front of the Main Building and facing it, is a copper statue of Christ with arms upraised with the legend "Venite Ad Me Omnes". Next to the Main Building is the Basilica of the Sacred Heart. Immediately behind the basilica is the Grotto, a Marian place of prayer and reflection. It is a replica of the grotto at Lourdes, France where the Virgin Mary reputedly appeared to Saint Bernadette Soubirous in 1858. At the end of the main drive (and in a direct line that connects through 3 statues and the Gold Dome), is a simple, modern stone statue of Mary.',
'answers': 'answer start': [515], 'text': ['Saint Bernadette Soubirous'],

### 4.1 Data preprocessing

For our main experiments that were not baselines, we randomly corrupt 5%, 10% and 15% of the training and validation dataset. We performed two types of corruption: word jumbling and word deletion. We perform this word jumbling and word deletion on the 'context' field of the randomly chosen samples. Only 5% of the 'context' is jumbled. SQUAD dataset from HuggingFace library comes with 'training' and 'validation' splits. Thus, we corrupted equal percentage of 'training' and 'validation' splits.

### 4.1.1 Word Jumbling:

In word jumbling function, we used a regular expression tokenizer from nltk library to tokenize the context of randomly chosen sample from the dataset into words. Using the length of the tokenized string, the function randomly picks 5% of total indices to corrupt. The words at the chosen indices are then jumbled. Finally, the function returns the re-joined sentence. **Example:** Input Context: Same as in word jumbling. Output Context: "Architecturally, the school has a Catholic character. Atop the Main Building's gold dome is a golden taetsu of the Virgin Mary"
"statue" has been jumbled.

### 4.1.2 Word Deletion:

In word deletion function, we used a regular expression tokenizer just like word jumbling to token the context. The function, instead of jumbling the randomly chosen indices, deletes the word from the sentence entirely. The function return the re-joined sentence without those chosen words.

**Example:** Input Context: "Architecturally, the school has a Catholic character. Atop the Main Building's gold dome is a golden statue of the Virgin Mary"
Output Context: "the school has a Catholic character. Atop the Main Building's gold dome is a golden statue of the Virgin Mary"
"Architecturally," has been deleted.

## 5 Baselines

### 5.1 Models Used:

We use four state of arts NLP models in our project, one for each of us to experiment with.

The first model is the BERT base, BERT is designed to pre-train deep bidirectional representations from the unlabeled text by jointly conditioning on both left and right context in all layers. It can be fine-tuned with just one additional output layer to create state-of-the-art models for a wide range of tasks like Question Answering. BERT base has 12 layers (i.e., Transformer blocks), 768 hidden sizes, and 12 self-attention heads. Total Parameters are 110M.

Another model is RoBERTa (Liu et al., 2019), which is based on BERT with better training. Specifically, RoBERTa is trained with dynamic masking (generate the masking pattern every time when feeding a sequence to the model), FULL-SENTENCES without Next Sentence Prediction (NSP) loss, large mini-batches, and a larger byte-level Byte-Pair Encoding (BPE). It follows the BERT large architecture, 24 layers, 1024 hidden sizes, 16 self-attention heads. Total Parameters are

355M.

We also test ALBERT (Lan et al., 2019), which is lite BERT that requires low GPU/TPU. AL-BERT base only has 12M parameters, with 12 layers, 768 hidden sizes and 128 embeddings. There are three main contributions that ALBERT makes over the design choices of BERT. The first is the factorization of the embedding parameters, the second is proposing cross-layer parameter sharing to improve parameter efficiency, and third is using a sentence-order prediction (SOP) loss, which avoids topic prediction and instead focuses on modeling inter-sentence coherence.

The last model that we are using is DistilBERT (Sanh et al., 2019), a smaller, faster, cheaper, and lighter BERT. It reduces the size of a BERT model by 40 percent while retaining 97 percent of its language understanding capabilities and being 60 percent faster. It has the same general architecture as BERT. The token-type embeddings and the pooler are removed while the number of layers is reduced by a factor of 2.

### 5.2 Hyperparameters Chosen:

There were several hyperparameters present during our experiments. We ran all experiments for one **epoch** and we used 16 as our **batch size** during training and validation. We first performed a random search for **learning rate** on a small subset of data. We found two promising learning rates: 2e-5 and 7e-5. 7e-5 gave us slightly better result so it was used for all experiments.

### 5.3 Evaluation Metrics:

The evaluation metrics that we are using are ROUGE and Exact Match. In ROUGE scores, we have used three different types of metrics such as ROUGE-1, ROUGE-2, and ROUGE-L. ROUGE stands for Recall-oriented Understudy for Gisting Evaluation. ROUGE-1 measures the unigram performance of model output and reference, while ROUGE-2 measures bigram performance. ROUGE-L measures the longest common subsequence (LCS) between our model output and reference. This metric calculates its score by comparing an automatically produced summary with a reference summary. This reference summary is typically generated by the human and is used as the reference with which the model-generated summary is compared. We report the mean F-measures for ROUGE-1, ROUGE-2 and ROUGE-L, which are defined as follows:

$$Precision = \frac{correct}{output\ length},$$

$$Recall = \frac{correct}{reference\ length},$$

and finally

$$F-measure = \frac{Precision \times Recall}{(Precision + Recall)/2}.$$

The Exact Match metric measures the percentage of predictions that match any one of the ground truth answers exactly. The F1 score metric is a looser metric measure the average overlap between the prediction and ground truth answer (Rajpurkar et al., 2016).

### 5.4 Baseline Results:

Table 1 shows our baseline results. It can be seen from the table that the model RoBERTa gives the best results, it has an F1 score of 91.75, and also the Exact Match score is pretty high with 85.128, after RoBERTa, ALBERT scores are high and then the BERT-base, the least performing model out of the four chosen models is DistilBERT with an F1 score of 85.725 and an Exact Match score of 76.3859

## 6 Your approach

### 6.1 Libraries and Compute Resources:

We used Google Colab Pro for the additional GPU resources. This allowed us to experiment on large models like RoBERTa. The primary programming language used in this project is Python 3. The main library that we used was HuggingFace/transformers and HuggingFace/datasets where we obtained the pretrained models and SQUAD datasets. Additional libraries we used were HuggingFace/metrics for ROUGE score and SQUAD metrics. For data visualization, we used matplotlib and seaborn. For general array operations, we used NumPy and pandas.

### 6.2 Experiment Overview:

In our project, we are using the SQUAD dataset which we import from the transformers library and we get our pretrained models such as DistilBERT, RoBERTa, ALBERT and BERT-BASE from hugging face. We are using the SQUAD 1.1 dataset and it has different attributes such as ID, title, context, question, and answers. This dataset is then split into train, validation datasets. Then we use specific tokenizers from huggingface for each of these models. These tokenizers convert the input into tokens and also convert these tokens to their corresponding ID in the pretrained vocabulary. We use fast tokenizers imported from hugging face because of their additional functionalities. Unlike other tasks, we cannot truncate our sentences which are extremely long here because we never know what specific part of the sentence contains the relevant context needed to answer the question. To counter this issue, a long example that consists of the necessary input features is allowed but this would be shorter than the maximum allowed length.

| Model Name | F1 | Exact Match | ROUGE-1 | ROUGE-2 | ROUGE-L |
|---|---|---|---|---|---|
| BERT-base | 86.2827 | 78.0889 | 0.4308 | 0.2673 | 0.4303 |
| RoBERTa | 91.7583 | 85.1282 | 0.4406 | 0.2785 | 0.44 |
| ALBERT | 88.7207 | 81.5231 | 0.4346 | 0.2715 | 0.4342 |
| DistilBERT | 85.7251 | 76.3859 | 0.4264 | 0.262 | 0.4258 |

Table 1: Baseline Results

We also use a hyperparameter known as document stride to allow for overlap in the case where we lose important context due to splitting. This truncation is only applicable to context and does not happen with questions. After this step, the tokenizers give us several features that have a certain overlap as well as being bound by the maximum length. We would then go through these features to see which overlap or feature contains the required context. We would need to provide the start and end positions for the answers in the tokens, which can be done by mapping parts of the original context to the tokens. The fast tokenizers for each model help to do this by returning an offset mapping. The first token would be the CLS token which is encoded as (0,0) because this does not correspond to any answer and doesn't add value to our context. Another method of these fast tokenizers called as sequence id helps us to find the specific parts of the offset which correspond to the question and the answer. This method returns NONE for the special tokens and 0 if the token comes from the question and 1 if the token comes from the context. We also account for the case where the model expands padding from the left, here we switch the order of question and context accordingly.

If the question is unanswerable based on the context we have in our feature, we set the CLS token to be both the start and the end. This can happen if the relevant context is present in another feature since SQUAD 1 does not have unanswerable questions. Now that we have prepossessed our data, we train with a batch size of 16 and a learning rate of 7e-05. Each epoch took us around 90 to 120 minutes for our models. We have evaluated the performance of our model using three metrics which are ROUGE scores, Exact Match and F1 score.

### 6.3 Results and Discussion

We have represented our results using nine tables and four figures. Each table consists of the performance of our four models upon various metrics such as F1, Exact Match, ROUGE-1, ROUGE-2 and ROUGE-L. We have used a learning rate of 7e-05 for all of our experiments with a batch size of 16. We have introduced noise into our data in two ways which are word jumbling and word deletion. We have introduced noise at levels of 5, 10 and 15 percent. Table 1 describes our baseline results upon the various metrics we used. We performed the baseline experiment upon our models with zero noise of any kind. We can see that RoBERTa performs the best across all the metrics. ALBERT comes a close second followed by BERT-base and DistilBERT. The pattern among other models also follows similarly with their being a clear demarcation of the performance of models. The same pattern continued across metrics for all the models.

Table 2 to Table 5 represents the performance of our models when the words are jumbled at different levels of noise such as 5, 10 and 15 percent. In metrics such as F1 score and Exact Match, we can see a steady performance decrease across models as the percentage of words being jumbled increases from 5 percent to 10 percent. We can observe the same patterns as baseline with RoBERTa still being the strongest and DistilBERT being the weakest across these metrics. In ROUGE scores, we observe a lesser degradation of performance compared to F1 and Exact Match across metrics. For the different ROUGE metrics, we can see an interesting observation with RoBERTa as the performance improves on a minuscule level when the word jumbling level is increased from 10 to 15 percent. This means that the model is not as susceptible to performance degradation in ROUGE as it is with non-ROUGE metrics. Among all other metrics, we can observe a degradation in performance corresponding with increasing noise levels. If we analyze on a metric level, we can observe that ROUGE typically suffers from minute performance degradation compared to other metrics. We can observe that RoBerta, which is the best performing model has the smallest degradation and DistilBERT, which is our worst performing model suffers from the most degradation. We can understand from this that we can achieve better results by going for a model with a strong baseline rather than fine tuning a model with a poor baseline to avoid performance degradation.

Table 6 to Table 9 represents the performance of our models when words are deleted at different levels such a 5,10 and 15 percent. In the case of word deletion, we can see that there has been a clear pattern of performance degradation across ROUGE and non-ROUGE metrics, The degradation of performance is also a bit more pronounced, especially in the case of exact match. Unlike the case of word jumbling, the performance degradation when noise levels are increased is not lesser for RoBERTa, the model with the best baseline. So we can understand that for word deletion, performance degradation is not negatively correlated with better baseline performance. But it is not truly random either. In fact, we can observe that DistilBERT, the model with the worst baseline suffers lesser performance degradation than any other model. This can also be observed across metrics although it is more pronounced in non-ROUGE metrics such as Exact Match and F1. Although, the rate of performance degradation is lowest for DistilBERT, it ends up being the worst performing model for 15 percent word deletion. This proves that although a strong baseline might not be as much of a factor as in word jumbling, it still has a strong impact on performance in the case of word deletion too.

We also have 4 figures with 8 different line plots to represent the performance of our models visually. We plot these graphs for the metrics F1 and Exact Match because those are the metrics for which we observe significant changes in performance. In Figure 1, we plot the results of word jumbling with respect to different levels of corruption as percentage. We can see that for both the F1 and Exact Match metrics, there is a sizeable gulf in the performances for RoBERTa and DistilBERT in comparison to the other models. The performance of BERT-Base and ALBERT is very similar. This is also true in the case of word deletion. Figure 2 represents the performances of our models for the task of word deletion. We can see that for RoBERTa and DistilBERT, the performance trends are similar for both word jumbling and word deletion. For BERT-Base, we can see that both Exact Match and F1 score metrics, do not suffer from linear performance degradation from 10 percent to 15 percent. So we can say that at higher noise levels, this model is more robust than the others. But for word deletion, the opposite is true. The rate of performance degradation is more in BERT-Base compared to the other models. So, we can understand that performance of BERT-Base is highly specific to the type of noise in our model.

The Figures 3 and 4 consist of line plots for the Word Jumbling and Word Deletion ratios. These ratios for each of our models are generated by comparison with the baseline performance. Since the performance of all the models has decreased with respect to the baseline after noise has been introduced, all of the ratios we obtain are less than 1. The line graphs for BERT-Base and ROBERTA are very similar for Exact Match, this means that the performance degradation in terms of exact match is very similar for both of them. In contrast, for F1 score, the performance degradation for BERT-Base is lower than the other models. DistilBERT has a linear performance degradation for Exact Match whereas it is performing a little better with regards to F1 score. RoBERTa is our best performing model but it is not the one with the least performance degradation, in both metrics and for both types of noises. For the F1 score metric, RoBERTa has the worst performance degradation amongst all the metrics.

# 7 Error analysis

Since we are focusing on comparing the performance of models with different levels of corrup-tions in the dataset, and due to the time constraint, we are not able to do detailed error analysis. However, much research has been done on the error analysis of Question-Answering System. For baseline models, common errors comes from failing to find the correct location of the answer in the context. With the corruption of the dataset, it is even harder for models to find the correct answer, which also explains why all models performs worse than their baseline.

We have analysed the some of the sentence where the model performs poorly, some of the points we have got after analyzing regarding the places where the model is going wrong includes when the question is not even related to the reference but still the model gives some answer, when the question contain some word and somewhere in the reference that word is used though it is not the answer, still the model relates these and gives that sentence or line as an answer though it is not the answer, then there are cases where there are multiple answers for the single question, the model only gives a single answer and not all the answers. There are also cases when there is correct answer in the give reference but the model leans towards the sentence where there are more overlapping word in the questions and the sentence.

We have experiment with the model on the Wikipedia Virat Kohli page with some sentences and we have received the following negative answers for the following question that we have asked where the model performed poorly.

Q: Where was virat born?
**A: 5 November 1988**

Q. What game does virat play?
**A: Royal Challengers Bangalore in the Indian Premier League**

Q. What awards did virat receive?
**A: Sir Garfield Sobers Trophy**

Q. What did virat receive?
**A: 26 million dollars**
**A: the man of the match**

Q: how many matches did virat kohli play?
**A: 213**

Q: Which name is also used to describe the Amazon rainforest in English?
**A: Hindustani**

## 8 Conclusion

None of the models that we experiment with is completely robust with context corruption, it means a robust model that can survive with the context corruption is needed. Due to the time restriction we are not be able to build such model but we believe more will be done in this area in the future.

We found that ROBERTa had the best baseline performance and even after introducing different levels of noise, it still managed to have a better performance than other models across metrics, But this does not mean that ROBERTa is the best model in terms of all aspects. It has a very steep performance degradation especially in metrics such as F1 score in particular. This means that at higher levels of noise, whether it be through word jumbling or word deletion or possibly other kinds of perturbations, this model might not be able to scale its performance. Models like DistilBERT had a much worse baseline but they proved to be more robust to noise. ALBERTA had a performance that was close to ROBERTa especially in the baseline but similarly, had a lower performance. In terms of ROUGE scores, a similar trend was followed but without any significant impact upon performance.

Based on all the different experiments we have conducted and the results we have obtained, we can understand that the baseline performance of a model is a very important component, especially for low levels of noise. The lack of robustness to noise of a model can be offset by a model with a sufficiently good baseline. However, this might not work when we introduce higher levels of noise. So based on the levels of noise we anticipate our text to have, we should make the appropriate choice of model. If we anticipate very high levels of noise and other human error, a model which is robust to noise and can suffer from lower performance degradation should be chosen. If we are operating at low levels of noise, we can go for a model with a stronger baseline. This decision could be taken because the gains made through baseline would not be completely offset by performance degradation at lower levels of noise.

We primarily found difficulty in training our models. If we were to continue with this line of inquiry that we introduced in the problem state-ment, we would look for better ways/resources to train the larger models that we couldn't use. Word jumbling and word dropping are simple forms of perturbations. So, we could explore other forms of perturbation that closer mimic human error and study their effects in a future study.

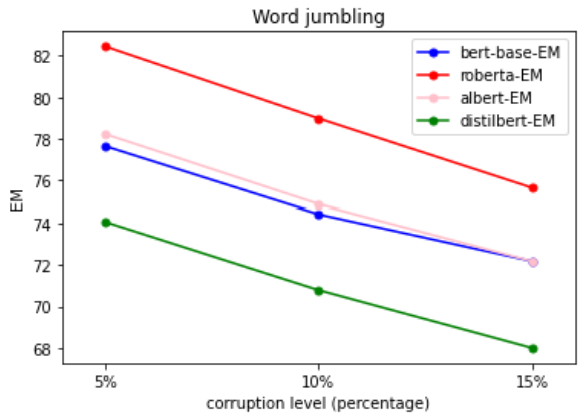
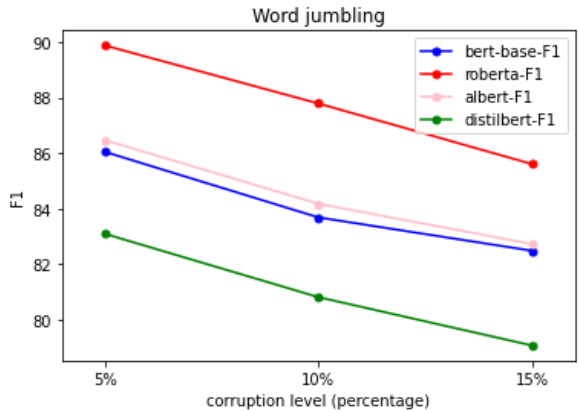

Figure 1: Word Jumbling results

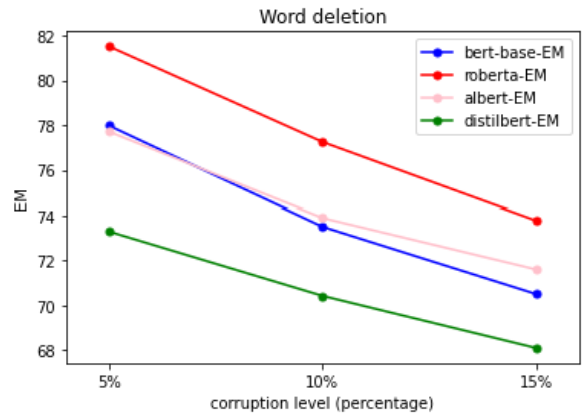
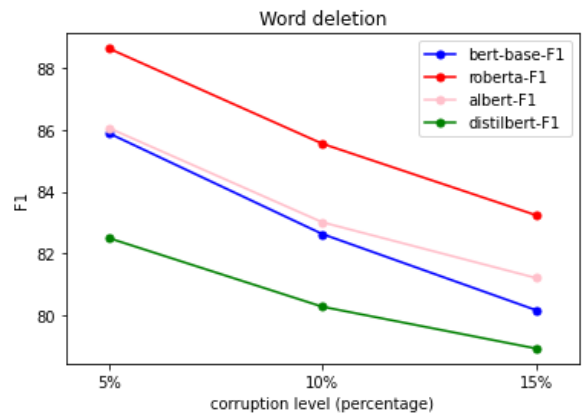

Figure 2: Word Deletion results

## 9 Contributions of group members

- Haoran Hu: wrote 25% of the report and did experiments of RoBERTa. Created graphs for the report.

- Vijaya Gajanan Buddhavarapu: wrote 25% of the report and did experiments on BERT. Created the word jumbling and word deletion functions.

- Sai Sameer Vennam: wrote 25% of the report and did experiments on ALBERT. Created tables in the report.

- Shashidhar Javaji Reddy: wrote 25% of the report and did experiments on DistilBERT. Created graphs for the report.

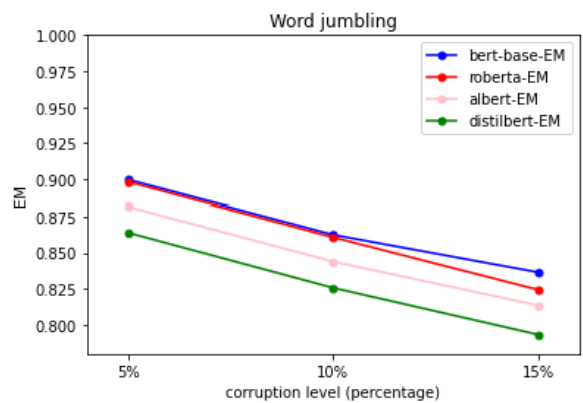
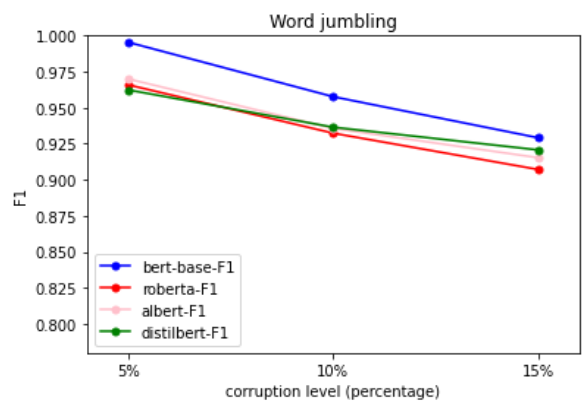

Figure 3: Word Jumbling Ratio results

| Corruption Level | F1 | Exact Match | ROUGE-1 | ROUGE-2 | ROUGE-L |
|---|---|---|---|---|---|
| 5% | 86.0547 | 77.6537 | 0.431 | 0.2671 | 0.4304 |
| 10% | 83.688 | 74.4181 | 0.4266 | 0.2619 | 0.4261 |
| 15% | 82.4879 | 72.1759 | 0.4256 | 0.2601 | 0.4252 |

Table 2: BERT-base: word jumbling

| Corruption Level | F1 | Exact Match | ROUGE-1 | ROUGE-2 | ROUGE-L |
|---|---|---|---|---|---|
| 5% | 89.8792 | 82.4414 | 0.4368 | 0.2737 | 0.4365 |
| 10% | 87.7883 | 78.9972 | 0.4291 | 0.2623 | 0.4285 |
| 15% | 85.6154 | 75.6481 | 0.4313 | 0.2658 | 0.4309 |

Table 3: RoBERTa: word jumbling

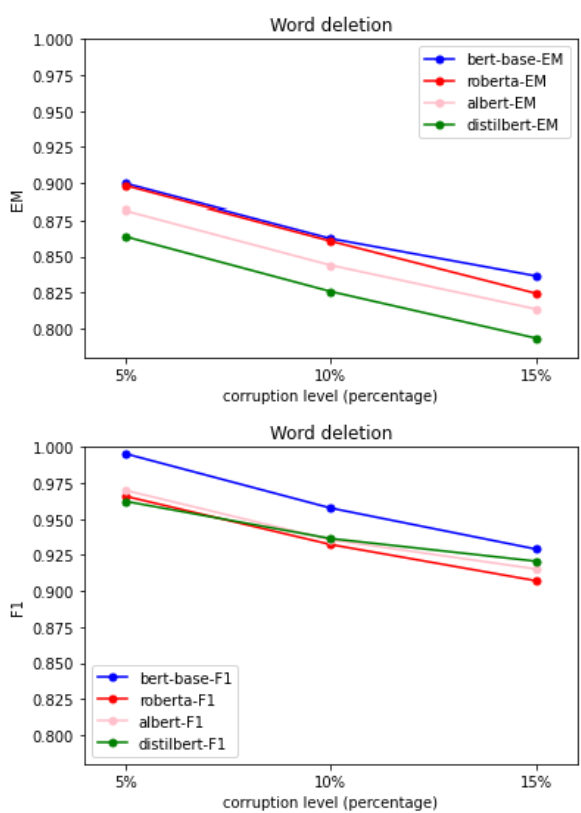

Figure 4: Word Deletion Ratio results

| Corruption Level | F1 | Exact Match | ROUGE-1 | ROUGE-2 | ROUGE-L |
|---|---|---|---|---|---|
| 5% | 86.4774 | 78.2308 | 0.4304 | 0.2660 | 0.4300 |
| 10% | 84.1842 | 74.8817 | 0.4276 | 0.2625 | 0.4271 |
| 15% | 82.7390 | 72.1854 | 0.4244 | 0.2584 | 0.4240 |

Table 4: ALBERT: word jumbling

| Corruption Level | F1 | Exact Match | ROUGE-1 | ROUGE-2 | ROUGE-L |
|---|---|---|---|---|---|
| 5% | 83.0008 | 74.0681 | 0.4242 | 0.2590 | 0.4236 |
| 10% | 80.8209 | 70.7947 | 0.4194 | 0.2534 | 0.4189 |
| 15% | 79.0863 | 68.0132 | 0.4166 | 0.2497 | 0.4161 |

Table 5: DistilBERT: word jumbling

| Corruption Level | F1 | Exact Match | ROUGE-1 | ROUGE-2 | ROUGE-L |
|---|---|---|---|---|---|
| 5% | 85.8922 | 77.9754 | 0.4305 | 0.2666 | 0.4300 |
| 10% | 82.6288 | 73.5099 | 0.4238 | 0.2590 | 0.4234 |
| 15% | 80.1732 | 70.5108 | 0.4204 | 0.2552 | 0.4197 |

Table 6: BERT-base: word deletion

| Corruption Level | F1 | Exact Match | ROUGE-1 | ROUGE-2 | ROUGE-L |
|---|---|---|---|---|---|
| 5% | 88.6330 | 81.5231 | 0.4344 | 0.2717 | 0.4340 |
| 10% | 85.5463 | 77.247 | 0.4293 | 0.2649 | 0.4289 |
| 15% | 83.2362 | 73.7840 | 0.4255 | 0.2602 | 0.4251 |

Table 7: RoBERTa: word deletion

| Corruption Level | F1 | Exact Match | ROUGE-1 | ROUGE-2 | ROUGE-L |
|---|---|---|---|---|---|
| 5% | 86.0629 | 77.7199 | 0.4304 | 0.2663 | 0.4300 |
| 10% | 83.0190 | 73.9072 | 0.4241 | 0.2590 | 0.4236 |
| 15% | 81.2135 | 71.6177 | 0.4216 | 0.2565 | 0.4211 |

Table 8: ALBERT: word deletion

| Corruption Level | F1 | Exact Match | ROUGE-1 | ROUGE-2 | ROUGE-L |
|---|---|---|---|---|---|
| 5% | 82.5054 | 73.3017 | 0.4234 | 0.2584 | 0.4228 |
| 10% | 80.2807 | 70.4351 | 0.4193 | 0.2537 | 0.4187 |
| 15% | 78.9387 | 68.1078 | 0.4158 | 0.2490 | 0.4153 |

Table 9: DistilBERT: word deletion